\documentclass[10pt,twocolumn,letterpaper]{article}

\usepackage{cvpr}
\usepackage{times}
\usepackage{epsfig}
\usepackage{eso-pic}
\usepackage{amssymb}

\usepackage[utf8]{inputenc} 
\usepackage[T1]{fontenc}    
\usepackage{url}            
\usepackage{booktabs}       
\usepackage{amsfonts}       
\usepackage{nicefrac}       
\usepackage{microtype}      
\usepackage{lipsum}
\usepackage{amsmath}
\usepackage{array}
\usepackage{float}
\usepackage{bm}
\usepackage{multirow}
\usepackage{booktabs}
\usepackage{subfigure}
\usepackage{algorithm}  
\usepackage{algorithmic}
\usepackage{graphicx}
\usepackage{xcolor}

\newcommand*{\ours}{Social-STGCNN }

\cvprfinalcopy 

\ifcvprfinal\fi

\begin{document}
\title{Social-STGCNN: A Social Spatio-Temporal Graph Convolutional Neural Network for Human Trajectory Prediction}

\author
{
Abduallah Mohamed\textsuperscript{1}, Kun Qian\textsuperscript{1}\\ Mohamed Elhoseiny\textsuperscript{2,3, **}, Christian Claudel\textsuperscript{1, **}\\ \textsuperscript{1}The University of Texas at Austin\hfill\textsuperscript{2}KAUST\hfill\textsuperscript{3}Stanford University \\ \small\{abduallah.mohamed,kunqian,christian.claudel\}@utexas.edu,  mohamed.elhoseiny@kaust.edu.sa  
}


\maketitle
\thispagestyle{empty}

\begin{abstract}
Better machine understanding of pedestrian behaviors enables faster progress in modeling interactions between agents such as autonomous vehicles and humans. Pedestrian trajectories are not only influenced by the pedestrian itself but also by interaction with surrounding objects. Previous methods modeled these interactions by using a variety of aggregation methods that integrate different learned pedestrians states. We propose the Social Spatio-Temporal Graph Convolutional Neural Network (Social-STGCNN), which substitutes the need of aggregation methods by modeling the interactions as a graph. Our results show an improvement over the state of art by 20\% on the Final Displacement Error (FDE) and an improvement on the Average Displacement Error (ADE) with 8.5 times less parameters and up to 48 times faster inference speed than previously reported methods. In addition, our model is data efficient, and exceeds previous state of the art on the ADE metric with only 20\% of the training data. We propose a kernel function to embed the social interactions between pedestrians within the adjacency matrix. Through qualitative analysis, we show that our model inherited social behaviors that can be expected between pedestrians trajectories. Code is available at \url{https://github.com/abduallahmohamed/Social-STGCNN}.
\end{abstract}

\section{Introduction}

\label{sec:intro}
\let\thefootnote\relax\footnotetext{ \textsuperscript{**} Equal advising.}

\begin{figure}[ht]
\begin{center}
\includegraphics[width=\linewidth]{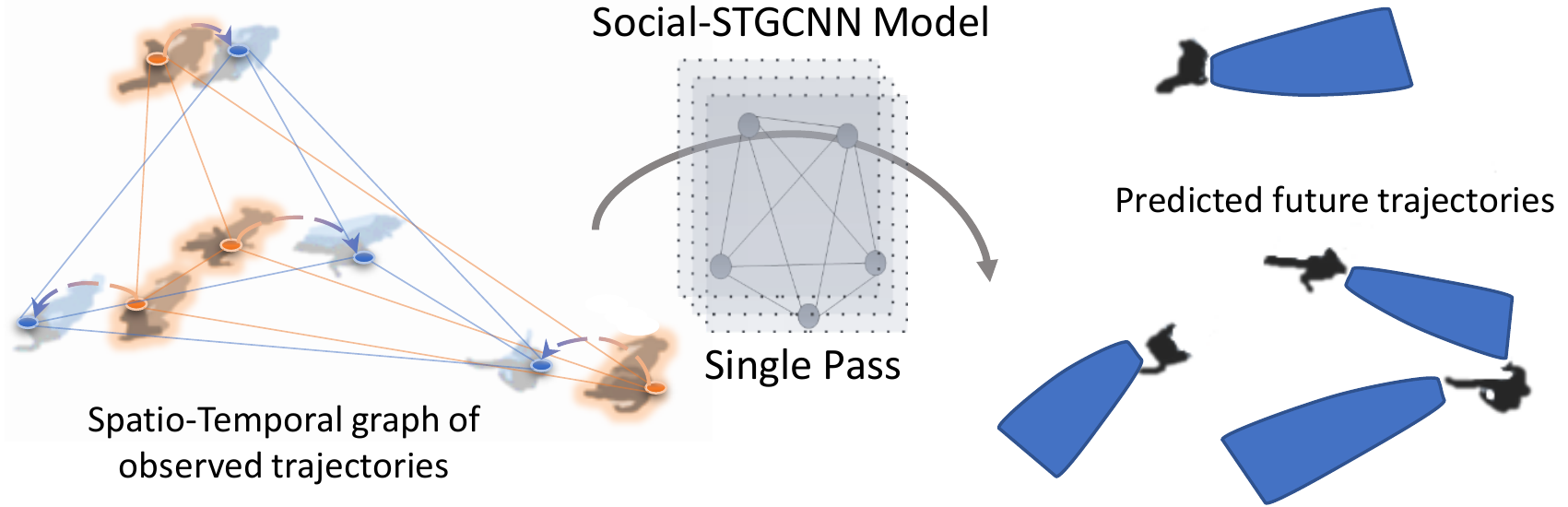}
   \caption {Pedestrian future trajectories prediction using the \ours model. The social interactions between pedestrians and their temporal dynamics are represented by a spatio-temporal graph. We predict the future trajectories in a single pass.}
   \end{center}
\label{fig:teaser}
\end{figure}

Predicting pedestrian trajectories is of major importance for several applications including autonomous driving and surveillance systems. In autonomous driving, an accurate prediction of pedestrians trajectories enables the controller to plan ahead the motion of the vehicle in an adversarial environment. For example, it is a critical component for collision avoidance systems or emergency braking systems~\cite{bai2015intention,morotomi2014collision,luo2018porca,raksincharoensak2016motion}. In surveillance systems, forecasting pedestrian trajectories is critical in helping identifying suspicious activities~\cite{luber2010people,yasuno2004pedestrian,musleh2010identifying}.

The trajectory of a pedestrian is challenging to predict, due to the complex interactions between the pedestrian with the environment. Objects potentially influencing the trajectory of a pedestrian include physical obstacles such as trees or roads, and moving objects including vehicles and other pedestrians. According to~\cite{moussaid2010walking}, 70\% of pedestrians tend to walk in groups. The interactions between pedestrians are mainly driven by common sense and social conventions. The complexity of pedestrian trajectory prediction comes from different social behaviors such as walking in parallel with others, within a group, collision avoidance and merging from different directions into a specific point. Another source of complexity is the randomness of the motion, given that the target destination and intended path of the pedestrian are unknown.

The social attributes of pedestrian motions encouraged researchers in this area to focus on inventing deep methods to model social interactions between pedestrians. In the Social-LSTM~\cite{alahi2016social} article, deep learning based model is applied to predict the pedestrians trajectories by modeling each pedestrian trajectory via a recurrent deep model. The outputs of recurrent models are made to interact with each other via a pooling layer. Several articles~\cite{manh2018scene,liang2019peeking,zhang2019sr} followed this direction. Social-LSTM~\cite{alahi2016social} modeled the pedestrian trajectories as a bi-variate Gaussian distribution, while some of others aimed at predicting deterministic trajectories. Another direction is to use Generative Adversarial Networks (GANs) for this task, assuming that the distribution of trajectories is multi-modal. Several articles~\cite{gupta2018social,sadeghian2019sophie,li2019conditional} used GANs to predict distributions of future trajectories. For these models, generators are designed using recurrent neural networks, and again, aggregation methods are relied upon to extract the social interactions between pedestrians. We argue that a limitation of earlier articles comes from the use of recurrent architectures, which are parameter inefficient and expensive in training~\cite{bai2018empirical}. We overcome this limitation through the use of temporal convolutional architectures.

In addition to the limitation of recurrent architectures, aggregation layers used in earlier works can also limit their performance. The aggregation layer takes the hidden states of the recurrent units as inputs. It is expected to assimilate a global representation of the scene, since each recurrent unit models a pedestrian trajectory. However, there are two issues within this type of aggregation. First, the aggregation in feature states is neither intuitive nor direct in modelling interactions between people, as the physical meaning of feature states is difficult to interpret. Second, since the aggregation mechanisms are usually based on heuristics like pooling, they could fail in modeling interactions between pedestrians correctly. For example, the pooling operation is known to be leaky in information~\cite{williams2018wavelet}. In order to directly capture the interactions between pedestrians and predict future paths from these, the recent article social-BiGAT~\cite{kosaraju2019social} relies on a graph representation to model social interactions. As the topology of graphs is a natural way to represent social interactions between pedestrians in a scene, we argue that it is a more direct, intuitive and efficient way to model pedestrians interactions than aggregation based methods. We also argue that social-BiGAT did not make the most of the graph representation, since they used it only as a pooling mechanism for recurrent units states. \ours benefits more from graph representation through modeling the scene with as spatio-temporal graph and performs on it.

We designed \ours to overcome the two aforementioned limitations. First, we model the pedestrians trajectories from the start as a spatio-temporal graph to replace the aggregation layers. The graph edges model the social interactions between the pedestrians. We propose a weighted adjacency matrix in which the kernel function quantitatively measure the influence between pedestrians. To address issues associated with recurrent units, our model manipulates over the spatio-temporal graph using a graph Convolutional Neural Networks (CNN)s and a temporal CNNs. This allows our model to predict the whole sequence in a single shot. Due to the above design, our model outperforms previous models in terms of prediction accuracy, parameters size, inference speed and data efficiency.


\section{Related work}
\label{sec:background}
The recent interest in autonomous driving has lead to increasing focus on pedestrian trajectory prediction. Recently, new deep models are making promising progresses on this task. In this section, we give a brief review of related work.

\noindent
\textbf{Human trajectory prediction using deep models}
Social-LSTM~\cite{alahi2016social} is one of the earliest deep model focusing on pedestrian trajectory prediction. Social-LSTM uses a recurrent network to model the motion of each pedestrian, then they aggregated the recurrent outputs using a pooling mechanism and predict the trajectory afterwards. Social-LSTM assumes the pedestrian trajectory follow a bi-variate Gaussian distribution, in which we follow this assumption in our model. Later works such as Peek Into The Future (PIF)~\cite{liang2019peeking} and State-Refinement LSTM (SR-LSTM)~\cite{zhang2019sr} extends~\cite{alahi2016social} with visual features and new pooling mechanisms to improve the prediction precision. It is noticeable that SR-LSTM~\cite{zhang2019sr} weighs the contribution of each pedestrian to others via a weighting mechanism. It is similar to the idea in Social-BiGAT~\cite{kosaraju2019social} which uses an attention mechanism to weigh the contribution of the recurrent states that represent the trajectories of pedestrians. Based on the assumption that pedestrian trajectories follow multi-modal distributions, Social-GAN~\cite{gupta2018social} extends Social LSTM~\cite{alahi2016social} into a Recurrent Neural Network (RNN) based generative model. Sophie~\cite{sadeghian2019sophie} used a CNNs to extract the features from the scene as a whole then a two way attention mechanism is used per pedestrian. Later on, Sophie concatenates the attention outputs with the visual CNN outputs then a Long Short Term Memory (LSTM) autoencoder based generative model is used to generate the future trajectories. The work CGNS~\cite{li2019conditional} is similar to Sophie~\cite{sadeghian2019sophie} in terms of the architecture but they used a Gated Recurrent Units(GRU)s instead of LSTMs. We notice that most previous works were circulating around two ideas, model each pedestrian motion using a recurrent net and combine the recurrent nets using a pooling mechanism. Recent work Social-BiGAT~\cite{kosaraju2019social} relies on graph attention networks to model the social interactions between pedestrians. The LSTM outputs are fed to the graph in Social-BiGAT. One key difference between our model \ours and Social-BiGAT is that we directly model pedestrian trajectories as a graph from the beginning, where we give meaningful values for vertices. 

\noindent
\textbf{Recent Advancements in Graph CNNs} Graph CNNs were introduced by~\cite{kipf2016semi} which extends the concept of CNNs into graphs. The Convolution operation defined over graphs is a weighted aggregation of target node attributes with the attributes of its neighbor nodes. It is similar to CNNs but the convolution operation is taken over the adjacency matrix of the graphs. The works~\cite{kipf2016variational, berg2017graph, schlichtkrull2018modeling} extend the graph CNNs to other applications such as matrix completion and Variational Auto Encoders. One of the development related to our work is the ST-GCNN~\cite{yan2018spatial}. ST-GCNN is a spatio-temporal Graph CNN that was originally designed to solve skeleton-based action recognition problem. Even though the architecture itself was designed to work on a classification task, we adapt it to suit our problem. In our work, ST-GCNNs extract both spatial and temporal information from the graph creating a suitable embedding. We then operate on this embedding to predict the trajectories of pedestrians. Details are shown in section~\ref{sec:model}.

\noindent
\textbf{Temporal Convolutional Neural Networks (TCNs)}
Starting from~\cite{bai2018empirical}, the argue between the usage of Recurrent Neural Networks (RNN)s versus the usage of temporal CNNs for sequential data modeling is highlighted. Introduced by~\cite{bai2018empirical}, Temporal Convolutional Neural Networks(TCNs) take a stacked sequential data as input and predict a sequence as a whole. This could alleviate the problem of error accumulating in sequential predictions made by RNNs. What is more, TCNs are smaller in size compared to RNNs. We were inspired by TCNs and designed a temporal CNN model that extends the capabilities of ST-GCNNs. More details about this are in the model description section~\ref{sec:model}.

\section{Problem Formulation}
\label{sec:formulation}
Given a set of $N$ pedestrians in a scene with their corresponding observed positions $tr_{o}^n, n \in \{1, \dots, N\}$ over a time period $T_{o}$, we need to predict the upcoming trajectories $tr_{p}^n$ over a future time horizon $T_{p}$. For a pedestrian $n$, we write the corresponding trajectory to be predicted as $tr^n_{p} = \{ \, \mathbf{p}_t^n = (\mathbf{x}_t^n, \mathbf{y}_t^n) \, |\, t \in \{1,\dots,T_{p}\} \}$, where $(\mathbf{x}_t^n, \mathbf{y}_t^n)$ are random variables describing the probability distribution of the location of pedestrian $n$ at time $t$, in the 2D space. We make the assumption that $(\mathbf{x}_t^n, \mathbf{y}_t^n)$ follows bi-variate Gaussian distribution such that $\mathbf{p}_t^n \sim \mathcal{N}(\mu_{t}^n ,\sigma_{t}^{n}, \rho_t^n)$. Besides, we denote the predicted trajectory as $\hat{\mathbf{p}}_t^n$ which follows the estimated bi-variate distribution $\mathcal{N}(\hat{\mu}_{t}^n ,\hat{\sigma}_{t}^{n}, \hat{\rho}_t^n)$. Our model is trained to minimize the negative log-likelihood, which defined as:
\begin{equation}
\label{eq:loss}
    L^n(\mathbf{W}) = - \sum_{t = 1}^{T_{p}} \log(\mathbb{P}((\mathbf{p}_t^n|\hat{\mu}_{t}^n,\hat{\sigma}_{t}^{n}, \hat{\rho}_t^n )) 
\end{equation}{}
in which $\mathbf{W}$ includes all the trainable parameters of the model, $\mu_{t}^n$ is the mean of the distribution,$\sigma_{t}^{n}$ is the variances and $\rho_t^n$ is the correlation.

\section{The \ours Model}

\label{sec:model}
\subsection{Model Description}
\begin{figure*}[ht]
\begin{center}
\includegraphics[width=\linewidth]{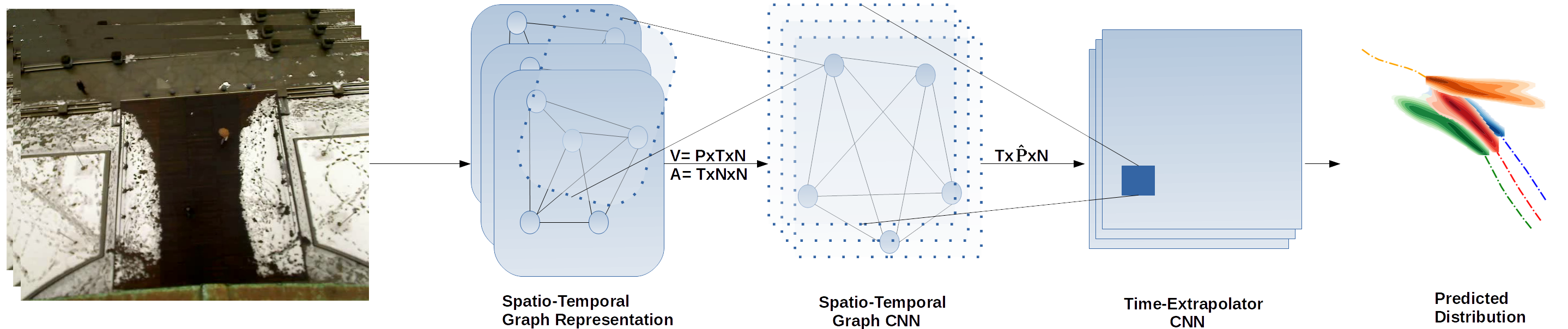}
\end{center}
   \caption{The \ours Model. Given $T$ frames, we construct the spatio-temporal graph representing $G=(V,A)$. Then G is forwarded through the Spatio-Temporal Graph Convolution Neural Networks (ST-GCNNs) creating a spatio-temporal embedding. Following this, the TXP-CNNs predicts future trajectories. $P$ is the dimension of pedestrian position, $N$ is the number of pedestrians, $T$ is the number of time steps and $\hat{P}$  is the dimensions of the embedding coming from ST-GCNN.}
\label{gr:modelpic}
\end{figure*}

The \ours model consists of two main parts: the Spatio-Temporal Graph Convolution Neural Network (ST-GCNN) and the Time-Extrapolator Convolution Neural Network (TXP-CNN). The ST-GCNN conducts spatio-temporal convolution operations on the graph representation of pedestrian trajectories to extract features. These features are a compact representation of the observed pedestrian trajectory history. TXP-CNN takes these features as inputs and predicts the future trajectories of all pedestrians as a whole. We use the name Time-Extrapolator because TXP-CNNs are expected to extrapolate future trajectories through convolution operation. Figure~\ref{gr:modelpic} illustrates the overview of the model.

\noindent
\textbf{Graph Representation of Pedestrian Trajectories} We first introduce the construction of the graph representation of pedestrian trajectories. We start by constructing a set of spatial graphs $G_t$ representing the relative locations of pedestrians in a scene at each time step $t$. $G_t$ is defined as $G_t = (V_t,E_t)$, where $V_t = \{v_t^i \; | \; \forall i \in \{1, \dots, N\} \}$ is the set of vertices of the graph $G_t$. The observed location $(x_t^i, y_t^i)$ is the attribute of $v_t^i$. $E_t$ is the set of edges within graph $G_t$ which is expressed as $E_t = \{ e_t^{ij} \; | \; \forall i,j \in \{1, \dots, N\} \}$. $e_t^{ij} = 1$ if $v_t^i$ and $v_t^j$ are connected, $e_t^{ij} = 0$ otherwise. In order to model how strongly two nodes could influence with each other, we attach a value $a_t^{ij}$, which is computed by some kernel function for each $e_t^{ij}$. $a_t^{ij}$s are organized into the weighted adjacency matrix $A_t$. We introduce $a_{sim,t}^{ij}$ as a kernel function to be used within the adjacency matrix $A_t$. $a_{sim,t}^{ij}$ is defined in equation~\ref{eq:ad_l2discon}. We discuss the details of $A_t$ kernel function later in section~\ref{sec:adjacency}.

\begin{equation}
\label{eq:ad_l2discon}
 a_{sim,t}^{ij} = 
  \begin{cases} 
       1/\lVert v^i_t-v^j_t\rVert_2 &\text{, } \lVert v^i_t-v^j_t\rVert_2 \neq 0 \\
      0& \text{, Otherwise.}
   \end{cases}
\end{equation}

\noindent
\textbf{Graph Convolution Neural Network} With the graph representation of pedestrian trajectories, we introduce the spatial convolution operation defined on graphs. For convolution operations defined on 2D grid maps or feature maps, the convolution operation is shown in equation~\ref{eq:emb}. 

\begin{equation}\label{eq:emb}
    z^{(l+1)} = \sigma(\sum_{h=1}^k \sum_{w=1}^k (p(z^{(l)},h,w)).\mathbf{w}^{(l)}(h,w)) \\
\end{equation}

where $k$ is the kernel size and $p(.)$ is the sampling function which aggregates the information of neighbors centering around $z$~\cite{dai2017deformable} and $\sigma$ is an activation function and $(l)$ indicates layer $l$. 

The graph convolution operation is defined as:
\begin{equation}\label{eq:graphconv}
    v^{i(l+1)} =\sigma(\frac{1}{\Omega} \sum_{v^{j(l)} \in B(v^{i(l)})}  p(v^{i(l)}, v^{j(l)}).\mathbf{w}(v^{i(l)}, v^{j(l)})) \\
\end{equation}
where $\frac{1}{\Omega}$ is a normalization term,  $B(v^i) = \{v^j |d(v^i, v^j) \leq D \}$ is the neighbor set of vertices $v^i$ and $d(v^i, v^j)$ denotes the shortest path connecting $v^i$  and $v^j$. Note that $\Omega$ is the cardinality of the neighbor set. Interested readers are referred to~\cite{kipf2016semi, yan2018spatial} for more detailed explanations and reasoning.

\noindent
\textbf{Spatio-Temporal Graph Convolution Neural Network (ST-GCNNs)}
ST-GCNNs extends spatial graph convolution to spatio-temporal graph convolution by defining a new graph $G$ whose attributes are the set of the attributes of $G_t$. $G$ incorporates the spatio-temporal information of pedestrian trajectories. It is worth noticing that the topology of $G_1, \dots, G_T$ is the same, while different attributes are assigned to $v_t^i$ when $t$ varies. Thus, we define $G$ as $(V, E)$, in which $V = \{ v^i \; | \; i\in \{1,\dots, N\} \}$ and $E = \{ e^{ij} \; | \; \forall i,j \in \{1, \dots, N\} \}$. The attributes of vertex $v^i$ in $G$ is the set of $v^i_t, \forall t \in \{0,\dots,T\}$. In addition, the weighted adjacency matrix $A$ corresponding to $G$ is the set of $\{A_1, \dots, A_T\}$. We denote the embedding resulting from ST-GCNN as $\bar{V}$. 

\noindent
\textbf{Time-Extrapolator Convolution Neural Network (TXP-CNN)}
The functionality of ST-GCNN is to extract spatio-temporal node embedding from the input graph. However, our objective is to predict further steps in the future. We also aim to be a stateless system and here where the TXP-CNN comes to play. TXP-CNN operates directly on the temporal dimension of the graph embedding $\bar{V}$ and expands it as a necessity for prediction. Because TXP-CNN depends on convolution operations on feature space, it is less in parameters size compared to recurrent units. A property to note regards TXP-CNN layer that it is not a permutation invariant as changes in the graph embedding right before TXP-CNN leads to different results. Other than this, if the order of pedestrians is permutated starting from the input to \ours then the predictions are invariant.

Overall, there are two main differences between \ours and ST-GCNN~\cite{yan2018spatial}. First, \ours constructs the graph in a totally different way from ST-GCNN with a novel kernel function. Second, beyond the spatio-temporal graph convolution layers, we added the flexibility in manipulating the time dimension using the TXP-CNN. ST-GCNN was originally designed for classification. By using TXP-CNN, our model was able to utilize the graph embedding originating from ST-GCNN to predict the futuree trajectories.

\subsection{Implementing \ours }
Several steps are necessary to implement the model correctly. We first normalize the adjacency matrix for the ease of learning. The adjacency matrix $A$ is a stack of $\{ A_1, \dots, A_T\}$, we symmetrically normalize each $A_t$ using the following form~\cite{kipf2016semi} \[A_t =  {\Lambda}_t^{-\frac{1}{2}} \hat{A_t} {\Lambda}_t^{-\frac{1}{2}}\]

where $\hat{A_t}=  A_t + {I}$ and ${\Lambda}_t$ is the diagonal node degree matrix of $\hat{A_t}$. We use $\hat{A}$ and $\Lambda$ to denote the stack of $\hat{A_t}$ and $\Lambda_t$ respectively. The normalization of adjacency is essential for the graph CNN to work properly, as outlined in~\cite{kipf2016semi}. 
We denote the vertices values at time step $t$ and network layer $l$ as $V_{t}^{(l)}$. Suppose $V^{(l)}$ is the stack of $V_{t}^{(l)}$. With the above definitions, we can now implement the ST-GCNN layers defined in equation~\ref{eq:graphconv} as follows.:
\begin{align}\label{eq:graph_conv_part}
	 f(V^{(l)}, A) = \sigma( {\Lambda}^{-\frac{1}{2}}\hat{A}{\Lambda}^{-\frac{1}{2}}V^{(l)}\mathbf{W}^{(l)})
\end{align}
where $\mathbf{W}^{(l)}$ is the matrix of trainable parameters at layer $l$.

After applying the ST-GCNN, we have features that compactly represent the graph. The TXP-CNN receives features $\bar{V}$ and treats the time dimension as feature channels. The TXP-CNN is made up of a series of residual connected CNNs. Only the first layer in TXP-CNN does not have a residual connection as it receives $\bar{V}$ from the ST-GCNNs, in which they differ in terms of the dimensions of the observed samples and the samples to be predicted.

\section{Datasets and Evaluation Metrics}
The model is trained on two human trajectory prediction datasets: ETH~\cite{pellegrini2009you} and UCY~\cite{lerner2007crowds}. ETH contains two scenes named ETH and HOTEL, while UCY contains three scenes named ZARA1, ZARA2 and UNIV. The trajectories in datasets are sampled every 0.4 seconds. Our method of training follows the same strategy as Social-LSTM~\cite{alahi2016social}. In Social-LSTM, the model was trained on a portion of a specific dataset and tested against the rest and validated versus the other four datasets. When being evaluated, the model observes the trajectory of 3.2 seconds which corresponds to 8 frames and predicts the trajectories for the next 4.8 seconds that are 12 frames.

Two metrics are used to evaluate model performance: the Average Displacement Error (ADE)~\cite{pellegrini2009you} defined in equation~\ref{eq:ADE} and the Final Displacement Error (FDE)~\cite{alahi2016social} defined in equation~\ref{eq:FDE}. Intuitively, ADE measures the average prediction performance along the trajectory, while the FDE considers only the prediction precision at the end points. Since \ours generates a bi-variate Gaussian distribution as the prediction, to compare a distribution with a certain target value, we follow the evaluation method used in Social-LSTM~\cite{alahi2016social} in which 20 samples are generated based on the predicted distribution. Then  the ADE and FDE are computed using the closest sample to the ground truth. This method of evaluation were adapted by several works such as Social-GAN~\cite{gupta2018social} and many more.

\begin{equation}
\label{eq:ADE}
    \text{ADE} = \frac {\sum\limits_{n \in N} \sum\limits_{t \in T_{p}} \lVert \hat{p}^n_{\text{t}}-p^n_{\text{t}} \rVert_2}{N \times T_{p}}
\end{equation}

\begin{equation}
\label{eq:FDE}
    \text{FDE} = \frac {\sum\limits_{n \in N} \lVert \hat{p}^n_{t}-p^n_{t} \rVert_2}{N} , t = T_{p}
\end{equation}

\section{Experiments and Results Analysis}
\label{sec:results}

\noindent
\textbf{Model configuration and training setup}
\ours is composed of a series of ST-GCNN layers followed by TXP-CNN layers. We use PReLU\cite{he2015delving} as the activation function $\sigma$ across our model. We set a training batch size of $128$ and the model was trained for $250$ epochs using Stochastic Gradient Descent (SGD). The initial learning rate is 0.01, and changed to 0.002 after 150 epochs. According to our ablation study in table~\ref{tb:ablation}, the best model to use has one ST-GCNN layer and five TXP-CNN layers. Furthermore, it is noticeable that when the number of ST-GCNN layers increases, the model performance decreases. Apparently, this problem of going deep using graph CNN was noticed by the work in~\cite{li2019deepgcns}, in which they proposed a method to solve it. Unfortunately, their solution does not extend to temporal graphs.

\begin{table}[ht]

\begin{tabular}{l||l|l|l|l}
  & 1           & 3           & 5           & 7           \\ \hline
 \midrule
1 & 0.47 / 0.78 & 0.47 / 0.84 & \textbf{0.44 / 0.75} & 0.48 / 0.87 \\ \hline
3 & 0.59 / 1.02 & 0.52 / 0.92 & 0.54 / 0.93 & 0.54 / 0.92 \\ \hline
5 & 0.62 / 1.07 & 0.57 / 0.98 & 0.59 / 1.02 & 0.59 / 0.98 \\ \hline
7 & 0.75 / 1.28 & 0.75 / 1.27 & 0.62 / 1.07 & 0.75 /1.28 \\
\midrule
\end{tabular}
\begin{center}

\caption{Ablation study of the \ours model. The first row corresponds to the number of TXP-CNN layers. The first column from the left corresponds to the number of ST-GCNN layers. We show the effect of different configurations of \ours on the ADE/FDE metric. The best setting is to use one layer for ST-GCNN and five layers for TXP-CNN.}
\end{center}
\label{tb:ablation}
\end{table}

\begin{table*}[ht]
\centering
\begin{tabular}{c||c|c|c|c|c|c}
 &  ETH & HOTEL & UNIV & ZARA1 & ZARA2 & AVG\\\hline
\midrule
Linear *~\cite{alahi2016social}  & 1.33 / 2.94 & 0.39 / 0.72 & 0.82 / 1.59 & 0.62 / 1.21 & 0.77 / 1.48 & 0.79 / 1.59\\\hline
SR-LSTM-2 * ~\cite{zhang2019sr}  & 0.63 / 1.25 & 0.37 / 0.74 & 0.51 / 1.10 & 0.41 / 0.90 & 0.32 / 0.70 & 0.45 / 0.94\\\hline
S-LSTM~\cite{alahi2016social}  & 1.09 / 2.35 & 0.79 / 1.76 & 0.67 / 1.40 & 0.47 / 1.00 & 0.56 / 1.17 & 0.72 / 1.54\\\hline
S-GAN-P~\cite{gupta2018social}  & 0.87 / 1.62 & 0.67 / 1.37 & 0.76 / 1.52 & 0.35 / 0.68 & 0.42 / 0.84 & 0.61 / 1.21\\\hline
SoPhie~\cite{sadeghian2019sophie}  & 0.70 / 1.43 & 0.76 / 1.67 & 0.54 / 1.24 & 0.30 / 0.63 & 0.38 / 0.78 & 0.54 / 1.15\\\hline
CGNS~\cite{li2019conditional}  & \textbf{0.62} / 1.40 & 0.70 / 0.93 & 0.48 / 1.22 & 0.32 / 0.59 & 0.35 / 0.71 & 0.49 / 0.97\\\hline
PIF~\cite{liang2019peeking} & 0.73 / 1.65 & \textbf{0.30} / \textbf{0.59} & 0.60 / 1.27 & 0.38 / 0.81 & 0.31 / 0.68 & 0.46 / 1.00\\\hline
STSGN~\cite{zhang2019stochastic}  & 0.75 / 1.63 & 0.63 / 1.01 & 0.48 / 1.08 & 0.30 / 0.65 & \textbf{0.26} / 0.57 & 0.48 / 0.99\\\hline
GAT~\cite{kosaraju2019social} & 0.68 / 1.29 & 0.68 / 1.40 & 0.57 / 1.29 &\textbf{ 0.29} / 0.60 & 0.37 / 0.75 & 0.52 / 1.07\\\hline
Social-BiGAT~\cite{kosaraju2019social} & 0.69 / 1.29 & 0.49 / 1.01 & 0.55 / 1.32 & 0.30 / 0.62 & 0.36 / 0.75 & 0.48 / 1.00\\\hline

\midrule
\textbf{\ours} &0.64 / \textbf{1.11} &0.49 / 0.85 &\textbf{0.44} / \textbf{0.79} & 0.34 / \textbf{0.53} &0.30 / \textbf{0.48} &\textbf{ \textbf{0.44}} / \textbf{0.75}\\\hline

\end{tabular}
\caption{ADE / FDE metrics for several methods compared to \ours are shown. The models with * mark are non-probabilistic. The rest of models used the best amongst 20 samples for evaluation. All models takes as an input 8 frames and predicts the next 12 frames. We notice that \ours have the best average error on both ADE and FDE metrics. The lower the better.}
\label{tb:results}

\end{table*}

\begin{table}[ht]

\begin{tabular}{l||l|l}
&Parameters count & Inference time              \\ \hline
\midrule
\multicolumn{1}{l||}{S-LSTM~\cite{alahi2016social}}    & 264K \textcolor{blue}{(35x)}      & \multicolumn{1}{l}{1.1789 \textcolor{blue}{(589x)}} \\ \hline
\multicolumn{1}{l||}{SR-LSTM-2~\cite{zhang2019sr}}   & 64.9K \textcolor{blue}{(8.5x)}     & \multicolumn{1}{l}{0.1578  \textcolor{blue}{(78.9x)}}  \\ \hline
\multicolumn{1}{l||}{S-GAN-P~\cite{gupta2018social}}   & 46.3K \textcolor{blue}{(6.1x)}      & \multicolumn{1}{l}{0.0968 \textcolor{blue}{(48.4x)}} \\ \hline
\multicolumn{1}{l||}{PIF~\cite{liang2019peeking}}       & 360.3K \textcolor{blue}{(47x)}     & \multicolumn{1}{l}{0.1145}  \textcolor{blue}{(57.3x)} \\ \hline
\midrule
\multicolumn{1}{l||}{\textbf{\ours}} & \textbf{7.6K}       & \multicolumn{1}{l}{\textbf{0.0020}} \\ \hline

\end{tabular}
\begin{center}
   \caption{Parameters size and inference time of different models compared to ours. The lower the better. Models were bench-marked using Nvidia GTX1080Ti GPU. The inference time is the average of several single inference steps. We notice that \ours has the least parameters size compared and the least inference time compared to others. The text in \textcolor{blue}{blue} show how many times our model is faster than others. } 
\end{center}

\label{tb:speed}
\end{table}

\subsection {Ablation Study of Kernel Function}
\label{sec:adjacency}
In this section, our objective is to find a suitable kernel function to construct the weighted adjacency matrix. The weighted adjacency matrix $A_t$ is a representation of the graph edges attributes. The kernel function maps attributes at $v^i_t$ and $v^j_t$ to a value $a^{ij}_t$ attached to $e^{ij}_t$. In the implementation of \ours, $A_t$ weights the vertices contributions to each other in the convolution operations. The kernel function can thus be considered as a prior knowledge about the social relations between pedestrians. A straightforward idea in designing the kernel function is to use the distance measured by the $L_2$ norm defined in equation~\ref{eq:ad_l2} between pedestrians to model their impacts to each other. However, this is against the intuition that the pedestrians tend to be influenced more by closer ones. To overcome this, we use similarity measure between the pedestrians. One of the proposals is to use the inverse of $L_2$ norm as defined in equation \ref{eq:ad_l2eps}. The $\epsilon$ term is added in denominator to ensure numerical stability. Another candidate function is the Gaussian Radial Basis Function~\cite{vert2004primer}, shown in equation~\ref{eq:ad_exp}. We compare the performance of these kernel functions through experiments. The case that all the values in adjacency matrix between different nodes are set to one is used as a baseline.

According to results listed in table~\ref{tb:adja}, the best performance comes from $a_{sim,t}^{ij}$ defined in function~\ref{eq:ad_l2discon}. The difference between functions~\ref{eq:ad_l2eps} and~\ref{eq:ad_l2discon} exists in the case where $\lVert v^i_t-v^j_t\rVert_2 = 0$. In function~\ref{eq:ad_l2discon}, we set $a_{sim,t}^{ij} = 0$ when $\lVert v^i_t-v^j_t\rVert_2 = 0$ because it is assumed that the two pedestrians can be viewed as the same person when they stay together. Without it, the model will have an ambiguity in the relationship between pedestrians. For this, we use $a_{sim,t}^{ij}$ in the definition of the adjacency matrix in all of our experiments. 


\begin{equation}
\label{eq:ad_l2}
a_{L_2,t}^{ij} = \lVert v^i_t-v^j_t\rVert_2
\end{equation}

\begin{equation}
\label{eq:ad_exp}
a_{exp,t}^{ij} = \frac{\exp{(-\lVert v^i_t-v^j_t\rVert_2)}}{\sigma}
\end{equation}

\begin{equation}
\label{eq:ad_l2eps}
a_{sim_\epsilon,t}^{ij} = \frac{1}{\lVert v^i_t-v^j_t\rVert_2+\epsilon}
\end{equation}

\begin{table}[ht]
\centering
\begin{tabular}{l|l}
                              Kernel function      & ADE / FDE                        \\ \hline
 \midrule
\multicolumn{1}{l|}{\(a_{L_2,t}^{ij}\)}                     & \multicolumn{1}{l}{0.48 / 0.84} \\ \hline
\multicolumn{1}{l|}{\(a_{exp,t}^{ij}\)}         & \multicolumn{1}{l}{0.50 / 0.84} \\ \hline
\multicolumn{1}{l|}{\(a_{sim_{\epsilon,t}}^{ij}\)}  & \multicolumn{1}{l}{0.48 / 0.88} \\ \hline
\multicolumn{1}{l|}{Just ones}                    & \multicolumn{1}{l}{0.49 / 0.79} \\ \hline
\midrule
\multicolumn{1}{l|}{\(a_{sim,t}^{ij}\)}                 & \multicolumn{1}{l}{\textbf{0.44 / 0.75}} \\ \hline
\end{tabular}
\begin{center}
    \caption{The effect of different kernel functions for the adjacency matrix $A_t$ over the \ours performance.}

\end{center}
\label{tb:adja}
\end{table}

\begin{figure*}[ht]
\begin{center}
\includegraphics[width=\linewidth]{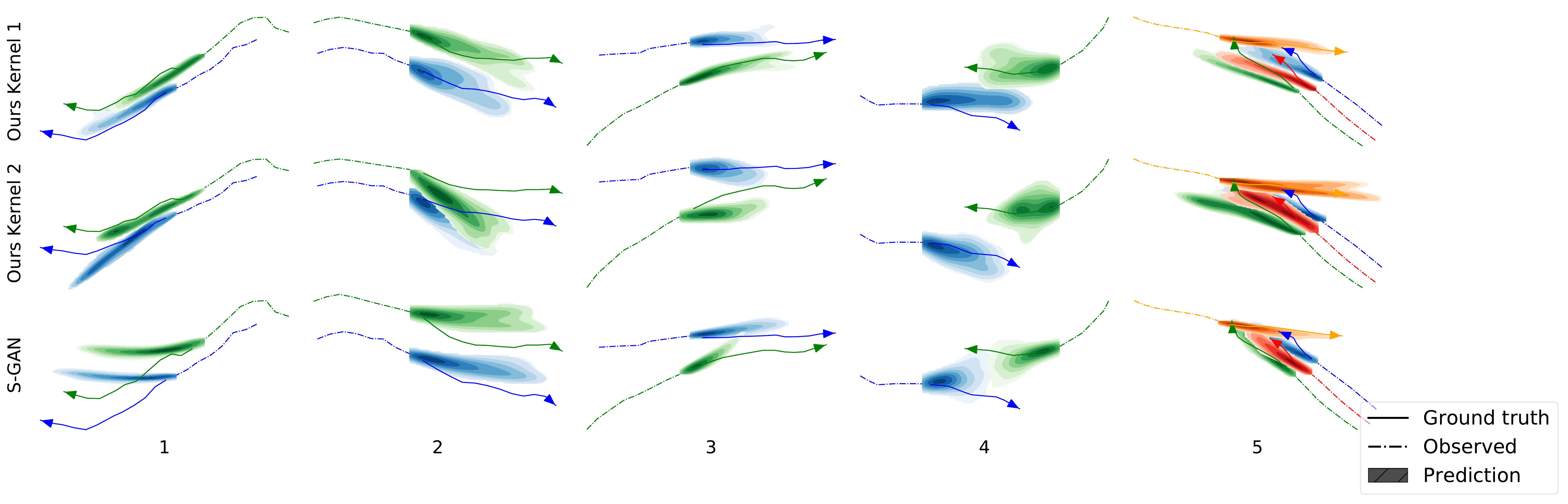}
\end{center}
   \caption{Qualitative analysis of \ours. We compare models trained with different kernel functions (Kernel 1: equation~\ref{eq:ad_l2} and Kernel 2: equation~\ref{eq:ad_l2discon}) versus previous models. Social-GAN \cite{gupta2018social} is taken as a baseline for the comparison. Illustration scenes are from the ETH~\cite{pellegrini2009you} and UCY~\cite{lerner2007crowds} datasets. We used the pre-trained Social-GAN model provided by \cite{gupta2018social}. A variety of scenarios are shown: two individuals walking in parallel (1)(2), two persons meeting from the same direction (3), two persons meeting from different directions (4) and one individual meeting another group of pedestrians from an angle (5). For each case, the dashed line is the true trajectory that the pedestrians are taking and the color density is the predicted trajectory distribution.}
\label{fig:qualana}
\end{figure*}
\begin{figure*}[!ht]
\begin{center}
\includegraphics[width=\linewidth]{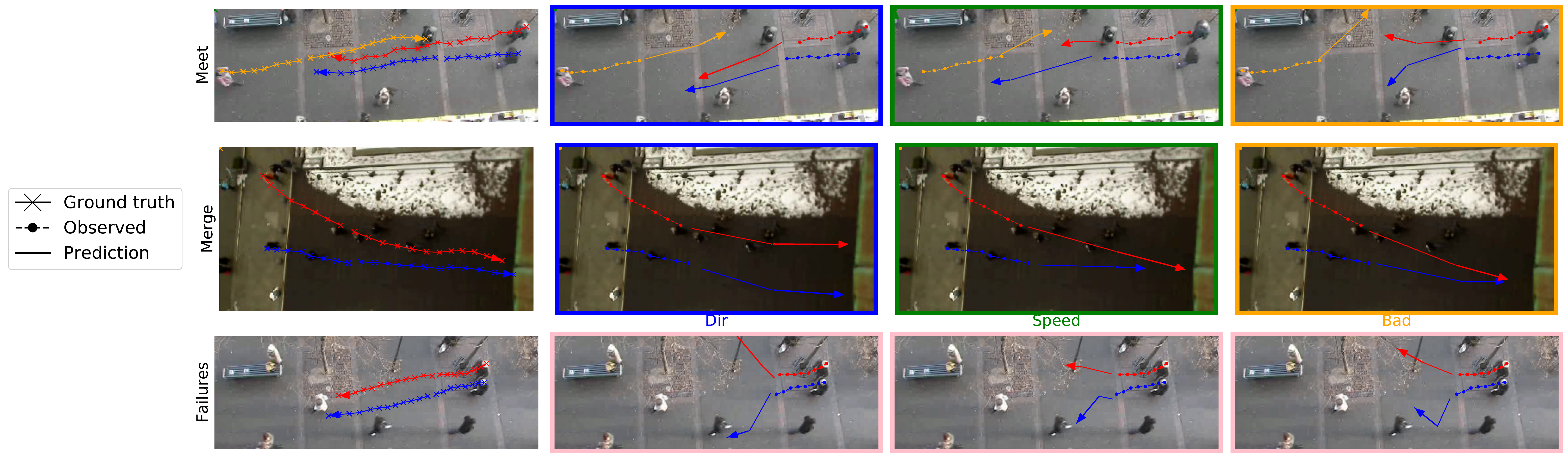}
\end{center}
   \caption{The first column is the ground truth, while the other columns illustrate samples from our model. The first two rows show two different scenarios where pedestrians merge into a direction or meet from opposite directions. The second and third columns show changes in \textcolor{green}{speed} or \textcolor{blue}{direction} in samples from our model. The last column shows \textcolor{orange}{undesired} behaviors. The last row show \textcolor{pink}{failed} samples.}
\label{fig:qualbkg}
\end{figure*}
\subsection{Quantitative Analysis}
The performance of \ours is compared with other models on ADE/FDE metrics in table~\ref{tb:results}. Overall, \ours outperforms all previous methods on the two metrics. The previous state of art on the FDE metric is SR-LSTM~\cite{zhang2019sr} with an error of 0.94. Our model has an error of 0.75 on the FDE metric which is about 20\% less than the state of the art. The results in qualitative analysis explains how \ours encourages social behaviors that enhanced the FDE metric. For the ADE metric, \ours is slightly better than the state-of-art SR-LSTM by 2\%. Also, it is better than the previous generative methods with an improvement ranging in between 63\% compared to S-LSTM~\cite{alahi2016social} and 4\% compared to PIF~\cite{liang2019peeking}. Interestingly, our model without the vision signal that contains scene context outperforms methods that utilized it such as SR-LSTM, PIF and Sophie.

\noindent
\textbf{Inference speed and model size} S-GAN-P~\cite{gupta2018social} previously had the smallest model size with 46.3k parameters. The size of \ours is 7.6K parameters only which is about one sixth of the number of parameters in S-GAN-P. In terms of inference speed, S-GAN-P was previously the fastest method with an inference time of 0.0968 seconds per inference step. The inference time of our model is 0.002 seconds per inference step which is about 48 $\times$ faster than S-GAN-P. Table~\ref{tb:speed} lists out the speed comparisons between our model and publicly available models which we could bench-mark against. We achieved these results because we overcame the two limitations of previous methods which used recurrent architecture and aggregation mechanisms via the design of our model.

\begin{figure}[ht]
\begin{center}
\includegraphics[width=\linewidth]{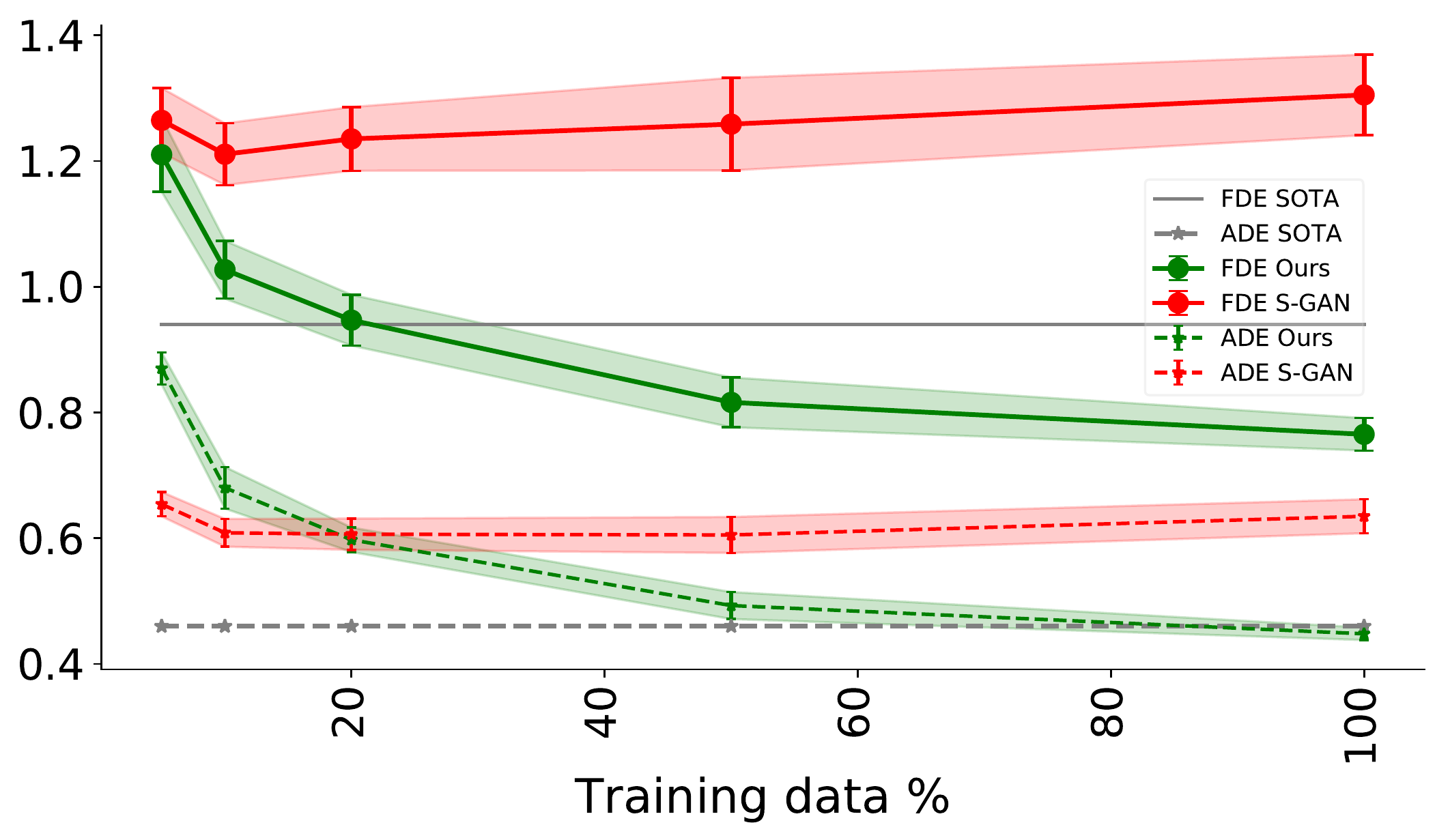}
   \caption{Model performance versus shrinked training dataset. The x-axis shows several randomly samples shrink percentages. The shade represents errors. The same shrinked data were used across the models. The figure shows our performance versus Social-GAN which is the closest model in terms of parameter size to ours.}
\end{center}

\label{fig:dataeff}
\end{figure}

\noindent
\textbf{Data Efficiency} In this section, we evaluate if the efficiency in model size leads to a better efficiency in learning from fewer samples of the data. We ran a series of experiments where 5\%, 10\%, 20\% and 50\% of the training data. The training data were randomly selected. Once selected, we fed the same data to train different models. Social-GAN is employed as a comparison baseline because it has least trainable parameters amongst previous deep models. Figure~\ref{fig:dataeff} shows the data learning efficiency experiments results with mean and error. We notice that our model exceeds the state of the art on the FDE metric when only 20\% of training data is used. Also, \ours exceeds the  performance of Social-GAN on the ADE metric when trained only on with 20\% of the training data. The results also show that S-GAN-P did not improve much in performance with more training data, unlike the present model. It is an interesting phenomenon that S-GAN-P does not absorb more training data. We assume that this behavior is due to the fact that GANs are data efficient because they can learn a distribution from few training samples. However, the training of GANs can easily fall into the problem of mode collapse. In comparison, the data efficiency of our model comes from the parameter efficiency.

\subsection{Qualitative Analysis}

The quantitative analysis section shows that \ours outperforms previous state-of-art in terms of ADE/FDE metrics. We now qualitatively analyze how \ours captures the social interactions between pedestrians and takes that into consideration when predicting the distributions. We show cases in which \ours successfully predicts collision free trajectories between pedestrians coming from different angles, maintains parallel walking, and correctly predicts the outcome of situations where person meets with a group of pedestrians. 

We qualitatively compare the prediction results between Social-GAN~\cite{gupta2018social}, \ours with $L_2$ norm (equation~\ref{eq:ad_l2}) as the kernel function and \ours with inverse $L_2$ norm (equation~\ref{eq:ad_l2discon}) as the kernel function.

\noindent
\textbf{Parallel walking } In scenarios one and two in figure~\ref{fig:qualana}, two pedestrians are walking in parallel. Usually, when people are walking in parallel they are tightly connected to each other and their momentum will be preserved in the future. The predictions by \ours and Social-GAN all show that these two pedestrians will keep walking in parallel in the future. However, the predicted density by \ours closely matches with the ground truth trajectory unlike the deviation we see in Social-GAN.

Using our proposed kernel function $a_{sim,t}$ defined in equation ~\ref{eq:ad_l2discon} for weighted adjacency matrix helps us model the social influences between pedestrians better than using the regular $L_2$ norm kernel function defined in equation~\ref{eq:ad_l2}. It is shown in scenes one and two that the model with $a_{sim,t}$ preforms much better in maintaining the relative location between people walking side by side. In scene five, similar behavior is observed.

\noindent
\textbf{Collision avoidance} Scenario three and Scenario four in figure~\ref{fig:qualana} are scenarios in which two pedestrians are heading towards similar or opposite directions. A collision could happen if they maintain their momentum. In scenario 3, two pedestrians are walking towards a similar direction. The forecast by Social-GAN acts linearly based on the momentum of the pedestrians and may lead to a collision. In the forecast of \ours, we notice that the trajectories are adjusted slightly such that they both avoid collision and align well with the observed momentum of pedestrians. As a result, \ours matches better with ground truth. In scenario four, Social-GAN fails to avoid the collision, while ours shows a realistic collision free path prediction.

\noindent
\textbf{Individual meeting a group} A more complex scenario is case five in figure~\ref{fig:qualana}, in which one person meets a group of parallel walking individuals. Our model suggests that the group of people still walk in parallel while adjusting their heading direction to avoid collision. In this case, although neither our model nor Social-GAN capture the ground truth trajectory very well, the predicted distribution by our model still makes sense from the social interaction point of view.

\noindent
\textbf{Diversity in samples and social behaviors} In order to understand in detail how \ours generates samples, we plot the samples generated from predicted bi-variate Gaussian distributions. There are two different scenarios in figure~\ref{fig:qualbkg}. In the first scene, three people meet from opposite directions. In the other scene, two people merge at an angle. Several patterns of samples could be generated by the predicted distributions. In column two in figure~\ref{fig:qualbkg}, the generated samples adjusts the advancing direction to avoid possible collisions in both scenes. Another social attribute of pedestrians is to slow down or accelerate to avoid crash. Samples in the third column in figure~\ref{fig:qualbkg} capture this attribute. This analysis shows that our samples encode different expected social behaviors of pedestrians. However, some samples show undesired behaviors such as collision or divergence in the last column. More cases of these undesired behaviors are in the last row of figure\ref{fig:qualbkg}.

\section{Conclusion}
\label{sec:conclusion}

In this article, we showed that a proper graph-based spatio-temporal setup for pedestrian trajectory prediction improves over previous methods on several key aspects, including prediction error, computational time and number of parameters. By applying a specific kernel function in the weighted adjacency matrix together with our model design, \ours outperforms state-of-art models over a number of publicly available datasets. We also showed that our configuration results in a data-efficient model and can learn from few data samples. We also qualitatively analyze the performance of \ours under situations such as collision avoidance, parallel walking and individual meeting a group. In these situations, \ours tend to provide more realistic path forecasts than several other reported methods. Furthermore, \ours is extremely efficient computationally, dividing the number of required parameters by a factor of 8.5, and boosting the inference speed by up to 48 $\times$ comparing to previous models. In the future, we intend to extend \ours to multi-modal settings that involve other moving objects including bicycles, cars and pedestrians.

\noindent
\textbf{Acknowledgement} The Authors would like to thank the reviewers for their suggestions which improved the paper. This research is supported by NSF (National Science Foundation) CPS No.1739964, CIS No.1636154 and CIS No.1917056. 

{\small
\bibliographystyle{ieee_fullname}
\bibliography{egbib}

\begin{thebibliography}{10}\itemsep=-1pt

\bibitem{alahi2016social}
Alexandre Alahi, Kratarth Goel, Vignesh Ramanathan, Alexandre Robicquet, Li
  Fei-Fei, and Silvio Savarese.
\newblock Social lstm: Human trajectory prediction in crowded spaces.
\newblock In {\em Proceedings of the IEEE conference on computer vision and
  pattern recognition}, pages 961--971, 2016.

\bibitem{bai2015intention}
Haoyu Bai, Shaojun Cai, Nan Ye, David Hsu, and Wee~Sun Lee.
\newblock Intention-aware online pomdp planning for autonomous driving in a
  crowd.
\newblock In {\em 2015 ieee international conference on robotics and automation
  (icra)}, pages 454--460. IEEE, 2015.

\bibitem{bai2018empirical}
Shaojie Bai, J~Zico Kolter, and Vladlen Koltun.
\newblock An empirical evaluation of generic convolutional and recurrent
  networks for sequence modeling.
\newblock {\em arXiv preprint arXiv:1803.01271}, 2018.

\bibitem{berg2017graph}
Rianne van~den Berg, Thomas~N Kipf, and Max Welling.
\newblock Graph convolutional matrix completion.
\newblock {\em arXiv preprint arXiv:1706.02263}, 2017.

\bibitem{dai2017deformable}
Jifeng Dai, Haozhi Qi, Yuwen Xiong, Yi Li, Guodong Zhang, Han Hu, and Yichen
  Wei.
\newblock Deformable convolutional networks.
\newblock In {\em Proceedings of the IEEE international conference on computer
  vision}, pages 764--773, 2017.

\bibitem{gupta2018social}
Agrim Gupta, Justin Johnson, Li Fei-Fei, Silvio Savarese, and Alexandre Alahi.
\newblock Social gan: Socially acceptable trajectories with generative
  adversarial networks.
\newblock In {\em Proceedings of the IEEE Conference on Computer Vision and
  Pattern Recognition}, pages 2255--2264, 2018.

\bibitem{he2015delving}
Kaiming He, Xiangyu Zhang, Shaoqing Ren, and Jian Sun.
\newblock Delving deep into rectifiers: Surpassing human-level performance on
  imagenet classification.
\newblock In {\em Proceedings of the IEEE international conference on computer
  vision}, pages 1026--1034, 2015.

\bibitem{kipf2016semi}
Thomas~N Kipf and Max Welling.
\newblock Semi-supervised classification with graph convolutional networks.
\newblock {\em arXiv preprint arXiv:1609.02907}, 2016.

\bibitem{kipf2016variational}
Thomas~N Kipf and Max Welling.
\newblock Variational graph auto-encoders.
\newblock {\em arXiv preprint arXiv:1611.07308}, 2016.

\bibitem{kosaraju2019social}
Vineet Kosaraju, Amir Sadeghian, Roberto Mart{\'\i}n-Mart{\'\i}n, Ian Reid,
  S~Hamid Rezatofighi, and Silvio Savarese.
\newblock Social-bigat: Multimodal trajectory forecasting using bicycle-gan and
  graph attention networks.
\newblock {\em arXiv preprint arXiv:1907.03395}, 2019.

\bibitem{lerner2007crowds}
Alon Lerner, Yiorgos Chrysanthou, and Dani Lischinski.
\newblock Crowds by example.
\newblock In {\em Computer graphics forum}, volume~26, pages 655--664. Wiley
  Online Library, 2007.

\bibitem{li2019deepgcns}
Guohao Li, Matthias Muller, Ali Thabet, and Bernard Ghanem.
\newblock Deepgcns: Can gcns go as deep as cnns?
\newblock In {\em Proceedings of the IEEE International Conference on Computer
  Vision}, pages 9267--9276, 2019.

\bibitem{li2019conditional}
Jiachen Li, Hengbo Ma, and Masayoshi Tomizuka.
\newblock Conditional generative neural system for probabilistic trajectory
  prediction.
\newblock {\em arXiv preprint arXiv:1905.01631}, 2019.

\bibitem{liang2019peeking}
Junwei Liang, Lu Jiang, Juan~Carlos Niebles, Alexander~G Hauptmann, and Li
  Fei-Fei.
\newblock Peeking into the future: Predicting future person activities and
  locations in videos.
\newblock In {\em Proceedings of the IEEE Conference on Computer Vision and
  Pattern Recognition}, pages 5725--5734, 2019.

\bibitem{luber2010people}
Matthias Luber, Johannes~A Stork, Gian~Diego Tipaldi, and Kai~O Arras.
\newblock People tracking with human motion predictions from social forces.
\newblock In {\em 2010 IEEE International Conference on Robotics and
  Automation}, pages 464--469. IEEE, 2010.

\bibitem{luo2018porca}
Yuanfu Luo, Panpan Cai, Aniket Bera, David Hsu, Wee~Sun Lee, and Dinesh
  Manocha.
\newblock Porca: Modeling and planning for autonomous driving among many
  pedestrians.
\newblock {\em IEEE Robotics and Automation Letters}, 3(4):3418--3425, 2018.

\bibitem{manh2018scene}
Huynh Manh and Gita Alaghband.
\newblock Scene-lstm: A model for human trajectory prediction.
\newblock {\em arXiv preprint arXiv:1808.04018}, 2018.

\bibitem{morotomi2014collision}
Kohei Morotomi, Masayuki Katoh, and Hideaki Hayashi.
\newblock Collision position predicting device, Sept.~30 2014.
\newblock US Patent 8,849,558.

\bibitem{moussaid2010walking}
Mehdi Moussa{\"\i}d, Niriaska Perozo, Simon Garnier, Dirk Helbing, and Guy
  Theraulaz.
\newblock The walking behaviour of pedestrian social groups and its impact on
  crowd dynamics.
\newblock {\em PloS one}, 5(4):e10047, 2010.

\bibitem{musleh2010identifying}
Basam Musleh, Fernando Garc{\'\i}a, Javier Otamendi,
  Jos{\'e}~M{\textordfeminine} Armingol, and Arturo De~la Escalera.
\newblock Identifying and tracking pedestrians based on sensor fusion and
  motion stability predictions.
\newblock {\em Sensors}, 10(9):8028--8053, 2010.

\bibitem{pellegrini2009you}
Stefano Pellegrini, Andreas Ess, Konrad Schindler, and Luc Van~Gool.
\newblock You'll never walk alone: Modeling social behavior for multi-target
  tracking.
\newblock In {\em 2009 IEEE 12th International Conference on Computer Vision},
  pages 261--268. IEEE, 2009.

\bibitem{raksincharoensak2016motion}
Pongsathorn Raksincharoensak, Takahiro Hasegawa, and Masao Nagai.
\newblock Motion planning and control of autonomous driving intelligence system
  based on risk potential optimization framework.
\newblock {\em International Journal of Automotive Engineering},
  7(AVEC14):53--60, 2016.

\bibitem{sadeghian2019sophie}
Amir Sadeghian, Vineet Kosaraju, Ali Sadeghian, Noriaki Hirose, Hamid
  Rezatofighi, and Silvio Savarese.
\newblock Sophie: An attentive gan for predicting paths compliant to social and
  physical constraints.
\newblock In {\em Proceedings of the IEEE Conference on Computer Vision and
  Pattern Recognition}, pages 1349--1358, 2019.

\bibitem{schlichtkrull2018modeling}
Michael Schlichtkrull, Thomas~N Kipf, Peter Bloem, Rianne Van Den~Berg, Ivan
  Titov, and Max Welling.
\newblock Modeling relational data with graph convolutional networks.
\newblock In {\em European Semantic Web Conference}, pages 593--607. Springer,
  2018.

\bibitem{vert2004primer}
Jean-Philippe Vert, Koji Tsuda, and Bernhard Sch{\"o}lkopf.
\newblock A primer on kernel methods.
\newblock {\em Kernel methods in computational biology}, 47:35--70, 2004.

\bibitem{williams2018wavelet}
Travis Williams and Robert Li.
\newblock Wavelet pooling for convolutional neural networks.
\newblock In {\em International Conference on Learning Representations}, 2018.

\bibitem{yan2018spatial}
Sijie Yan, Yuanjun Xiong, and Dahua Lin.
\newblock Spatial temporal graph convolutional networks for skeleton-based
  action recognition.
\newblock In {\em Thirty-Second AAAI Conference on Artificial Intelligence},
  2018.

\bibitem{yasuno2004pedestrian}
Masahiro Yasuno, Noboru Yasuda, and Masayoshi Aoki.
\newblock Pedestrian detection and tracking in far infrared images.
\newblock In {\em 2004 Conference on Computer Vision and Pattern Recognition
  Workshop}, pages 125--125. IEEE, 2004.

\bibitem{zhang2019stochastic}
Lidan Zhang, Qi She, and Ping Guo.
\newblock Stochastic trajectory prediction with social graph network.
\newblock {\em arXiv preprint arXiv:1907.10233}, 2019.

\bibitem{zhang2019sr}
Pu Zhang, Wanli Ouyang, Pengfei Zhang, Jianru Xue, and Nanning Zheng.
\newblock Sr-lstm: State refinement for lstm towards pedestrian trajectory
  prediction.
\newblock In {\em Proceedings of the IEEE Conference on Computer Vision and
  Pattern Recognition}, pages 12085--12094, 2019.

\end{thebibliography}
}

\onecolumn
\section*{Supplementary: More Qualitative Results}
\begin{figure*}[ht]
\centering
\includegraphics[width=\textwidth]{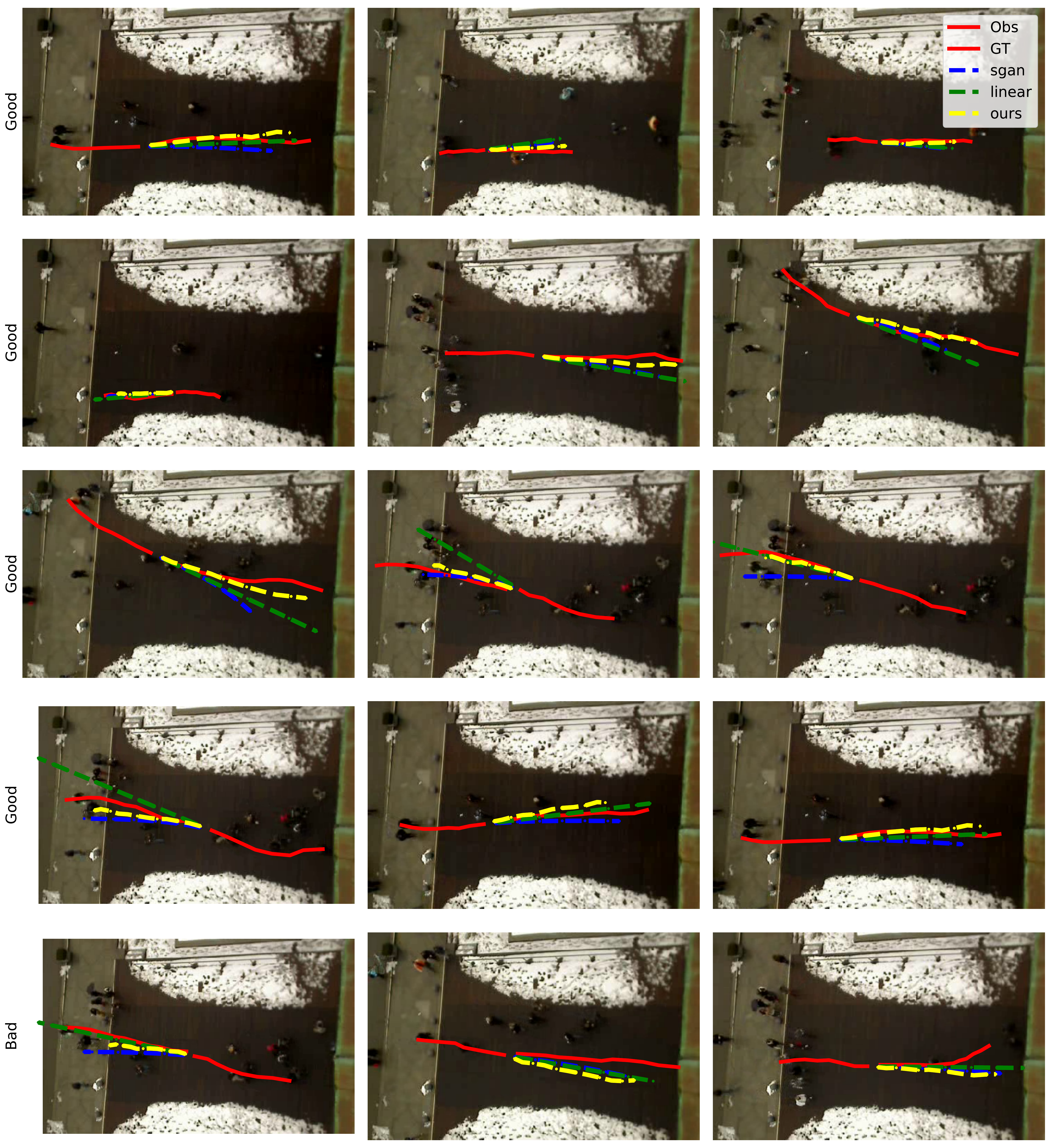}
\caption{Comparison of averaged trajectories predictions per model. Scenes were taken from the ETH dataset~\cite{pellegrini2009you}.}
\label{suppQ}
\end{figure*}

We compare our performance against linear model and Social-GAN~\cite{alahi2016social} predictions. Unlike the linear or Social-GAN predictions, in which they diverge and do not account for the variation in the pedestrian motion. \ours is able to track and align precisely with the ground truth. This can be seen in the first four rows in figure~\ref{suppQ}. The last row shows cases where our averaged trajectory fails.

\end{document}